\documentclass[a4paper]{article}

\usepackage{INTERSPEECH2018}
\usepackage[utf8]{inputenc}
\usepackage{tabularx}
\title{Do Autonomous Agents Benefit from Hearing?}

\name{Abraham Woubie, Anssi Kanervisto, Janne Karttunen, and Ville Hautamäki}
\address{
  School of Computing, University of Eastern Finland, Joensuu, Finland}
\email{\{abraham.zewoudie, anssk, jannkar, villeh\}@uef.fi}

\begin{document}

\maketitle
\begin{abstract}

Mapping states to actions in deep reinforcement learning is mainly based on visual information. The commonly used approach for dealing with visual information is to extract pixels from images and use them as state representation for reinforcement learning agent. But, any vision only agent is handicapped by not being able to sense audible cues. Using hearing, animals are able to sense targets that are outside of their visual range. In this work, we propose the use of audio as complementary information to visual only in state representation. We assess the impact of such multi-modal setup in reach-the-goal tasks in ViZDoom environment. Results show that the agent improves its behaviour when visual information is accompanied with audio features.
\end{abstract}

\section{Introduction}

The recent advances in deep learning makes it possible to extract high-level features from raw sensory data that lead to breakthroughs in computer vision \cite{sallans2004reinforcement} and speech recognition \cite{graves2013speech}. Recently, Convolutional Neural Networks (CNNs) have shown a remarkable progress in visual recognition tasks \cite{russakovsky2015imagenet} and audio tagging \cite {lederle2018combining}. They have also successfully been applied for feature extraction and classification tasks in the audio domain \cite{abdel2014convolutional,piczak2015environmental}. 

The design and training of an agent that interacts with an environment and solves a given problem has been studied for years using different methods including reinforcement learning \cite{sutton1998introduction}. Reinforcement learning studies problems and algorithms that learn policies so that the decisions maximize long term reward from the environment \cite{sutton1998introduction,kaelbling1996reinforcement}. 
Reinforcement learning has been successfully used in different tasks such as object recognition \cite{ba2014multiple} and solving physics-based control problems \cite {heess2015learning}. The recent advances in deep reinforcement learning have allowed autonomous agents to perform well on different games using only visual information \cite{vinyals2017starcraft}. It has also been applied successfully in different applications including Atari \cite {mnih2015human}, robotic arms \cite{levine2016end} and self-driving vehicles \cite{shalev2016safe}. However, only visual information is used as state representation in all of these mentioned and other related works \cite{mnih2015human}. 

\begin{figure}[t]
\centering
\includegraphics[width=\columnwidth]{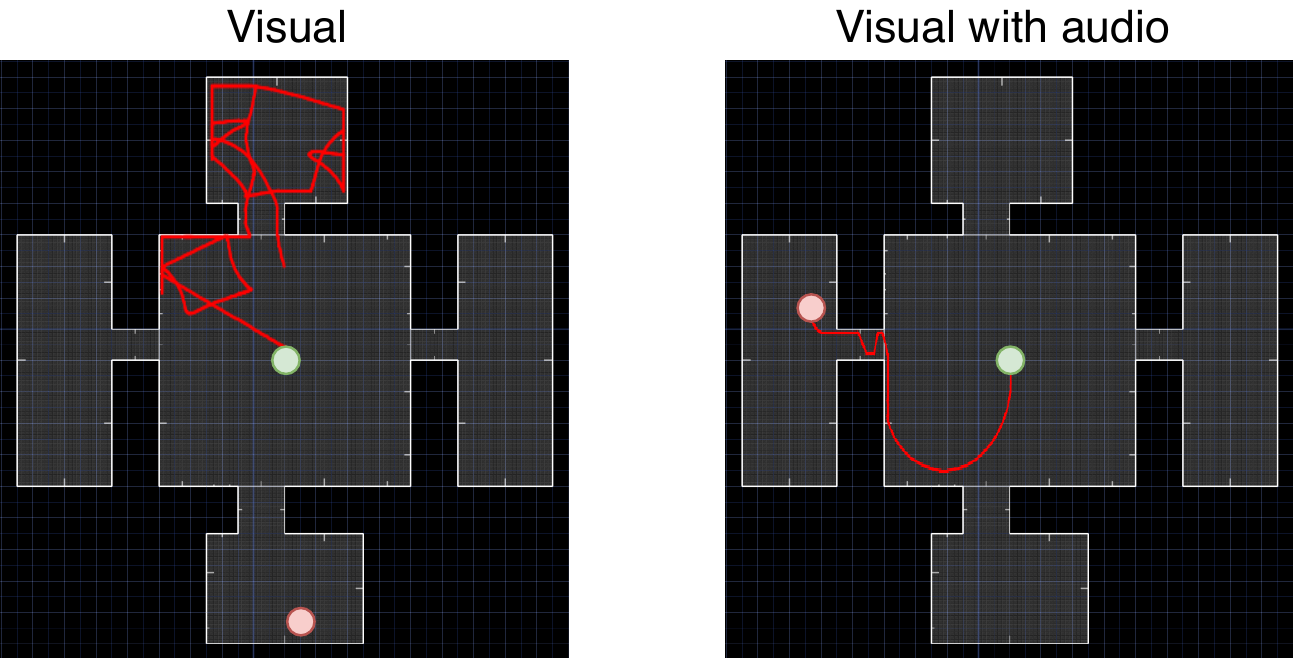}
\caption{ \it The agent (i.e., green) searching for the goal (i.e., red). By providing agent with sound from the goal, agent was able to reach the goal more reliably when trained with reinforcement learning.}
\label{fig:graphical_abstract}
\end{figure}

Learning using only visual information may not always be easy for the learning agent.  For example, it is difficult for the agent to reach the target using only visual information in scenarios where there are many rooms and there is no direct line of sight between the agent and the target. Without recurrent neural networks or other architectures that allow memorizing the past, the agent can't recall which rooms it has already searched.\footnote{While the task of this work could be solved with recurrent neural networks, this work focuses on the benefits from using audio information rather than solving the task.}.  We see example of searching task in Fig. \ref{fig:graphical_abstract} where an agent with only visual information is not able to systematically search the goal. Thus, the use of audio features could provide valuable information for such problems. Hence, in this work, we propose the use of audio as complementary input information to visuals. The work assesses the impact of audio features using different environment setups. The visual and audio information are used together to find the best action for a given state. 

This work was motivated by the natural tendency of guiding with spoken language when assisting someone to try to find an object. This lead us to a question "Why hearing is not used for the autonomous agents? One way to give agent a hearing would be utilizing an automatic speech recognition component. However, using an automatic speech recognition component may not be ideal as it would not necessarily detect audio signals other than speech, e.g. whistling. Hence, we study about training the agent with direct access to raw audio. 

The following sets of experiments have been carried out on VizDoom environment \cite{Kempka2016ViZDoom}: using only visuals, and using visuals together with audio features. We also experiment the impact of using audio information occasionally and all time. Simulation results on VizDoom environment manifest that a Deep Q-Network trained using visual and audio information provides better average reward compared to using only visual. 

\section{Deep Q-Network} \label{DQN}
\textit{Deep Q-networks} (DQN) were proposed by \cite{mnih2015human}, making it possible to use Q-learning \cite{watkins1992q} in complex environments like video games. DQN uses deep learning to estimate the value function rather than a discrete table, and uses two techniques to make learning stable: replay memory and target networks. DQN has been successfully applied to play a set of Atari games \cite{mnih2015human}, and Doom \cite{Kempka2016ViZDoom}. Because of these reasons and its simplicity, DQN is used in our experiments.

Reinforcement learning \cite{sutton1998introduction} methods aim to solve a task in an environment by gathering experiences from the environment and learning from them. The environment is modelled as a Markov Decision Process which consists of a set of possible states $s \in S$, a set of possible actions $a \in A$, a reward function $R(s_t, a_t, t_{t+1}) \in \mathbb{R}$ and transition distribution between the states $P(s_{t+1} | s_t, a_t) \in [0,1]$. Values at different time steps are distinguished with $t \in \mathbb{N}$. In each state, a policy $\pi : s \rightarrow a$ selects an action $a_t$, after which the environment will move to next state $s_{t+1} \sim P(s_{t+1} | s_t, a_t)$. The policy (agent) then receives a reward $r_t = R(s_t, a_t, s_{t+1})$ based on the experience. For compactness, we focus on \textit{episodic games}, where the game ends once agent reaches a terminal state. 

The goal of reinforcement learning is to learn a policy that maximizes the return from any time step $t \in \mathbb{N}$ till end of the episode $T \in \mathbb{N}$
\begin{equation}
    G_t = \sum^T_{k=0} \gamma^k r_{t+k} ,
\end{equation}
where \textit{discount factor} $\gamma \in (0,1]$ is used to stabilize the learning and/or weight the importance of future states vs. nearby states. 

Q-learning \cite{watkins1992q} solves this by learning \textit{state-action value function} $Q : s \times a \rightarrow G$, which represents the expected return from taking an action $a_t$ in state $s_t$. If we knew the true state-action values, selecting action with highest value $a_t = \arg\max_a Q(s_t, a)$ equals to optimal policy \cite{watkins1992q}. Q-learning learns this function by creating a table of all possible state-action pairs, and updating the value on each experience with
\begin{equation}
 \begin{aligned}
 Q(s_t, a_t) &  =  \\ 
 & Q(s_t, a_t) +  \alpha \left(r_t + \gamma \max_a Q(s_{t+1}, a) - Q(s_t, a_t) \right)  
  \end{aligned}
  \label{eq:qupdate}
\end{equation}
where \textit{learning rate} $\alpha \in \mathbb{R}$ controls rate of learning.
DQN extends this by using a deep neural network rather than a table to estimate the $Q$-function. This allows the use of Q-learning with high-dimensional states and actions (e.g. video games, where states are pixel of an image). The update rule of DQN (Eq. \ref{eq:qupdate}) is modified for training neural networks. To stabilize learning, DQN uses a replay memory and target networks \cite{mnih2015human}.

To encourage trying out different solutions to the task, we use $\epsilon$-greedy exploration strategy, where the DQN selects random action with probability $\epsilon$ and optimal action with probability $1 - \epsilon$. This value $\epsilon$ starts high at the beginning of the training, and slowly anneals towards zero over training to encourage using more optimal actions instead of always taking random actions (exploitation). 

This work uses DQN agent as the basis for all experiments because of its flexibility to different types of data (i.e image pixels and audio features). 

\section{Multimodal Reinforcement Learning Agent} \label{multimodal}




This work proposes the use of audio information along with the visual information. For the experiments, we use two different type of features based on the audio information: Pitch of the audio and raw samples. As it is displayed in Fig. \ref{fig:proposed}, the autonomous agent is trained using two sources of input (i.e., pixels and audio features). 

\subsection{Pitch}
Pitch is the variations in fundamental frequency which serves as an important acoustic cue for tone, lexical stress, and intonation \cite{ververidis2006emotional}. The pitch values are extracted from the audio clip and fed into the CNN architecture.

\subsection{Raw Audio}
Raw audio features are the samples of the signal itself, normalized appropriately for the neural network for processing. Using samples directly removes need for feature extraction, but may be too complex data for learning algorithm to use appropriately.


\subsection{Proposed Reinforcement Learning System Architecture} \label{proposed_architecture}

\begin{figure}[t]
\centering
\includegraphics[width=\columnwidth]{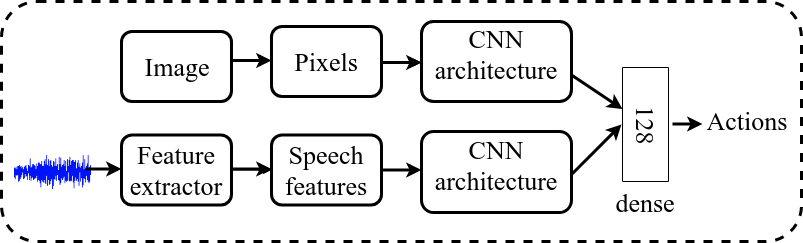}
\caption{ \it Proposed reinforcement learning architecture.}
\label{fig:proposed}
\end{figure}

Our proposed system uses both visual and audio features to train the agent. Both input sources have their own  convolutional layers. As it is displayed in Fig. \ref{fig:proposed}, the learned high-level features are combined in the final layer of the network by concatenating the outputs of the two separate CNN heads. 

In the experiments of this work, we use raw audio as audio for the agent. Specifically, we encode information about the environment (distance to the goal) into the pitch of the played speech sample. Then, the sample is provided to the agent along with the image. Note that the pitch of the audio sample is updated every time step. 

When using pitch features, we extract pitch for the sample with windowing of size $30$ms and step size of $10$ms, resulting into a vector of length $114$ elements for the neural network. Since distance to the goal is encoded in the overall pitch of the speech sample, these features could be easily digested for useful information for the agent (higher pitch equals closer to target). These features work as a sanity check that providing information on distance to goal is beneficial to the agent. 

For raw audio, we select $100$ speech samples for the network. We limit the number of samples to $100$ to reduce computational costs. Unlike with pitch features, agent now has to learn to process the frequency of signal provided to obtain the encoded information.

\section{Experimental Setup} \label{experiments}

We use VizDoom \cite{Kempka2016ViZDoom} as the environment in this work because of its flexibility, easiness to use, and efficient 3D platform. It allows developing artificial intelligence bots that play DOOM \cite{id1993doom} using visual information. The experiments have been carried out on machine with Intel Xeon processor, Nvidia RTX 2080ti, CUDA 10.0, Python 3.6, TensorFlow 1.12 and Ubuntu LTS 18.04.

\subsection{Scenario}

\begin{figure}[t]
  \centering
  \begin{minipage}[b]{0.2\textwidth}
    \includegraphics[width=\textwidth]{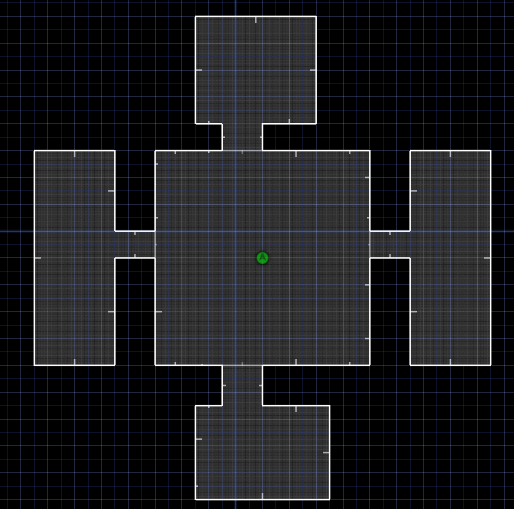}
  \end{minipage}
  \hfill
  \begin{minipage}[b]{0.265\textwidth}
    \includegraphics[scale=0.36,width=\textwidth]{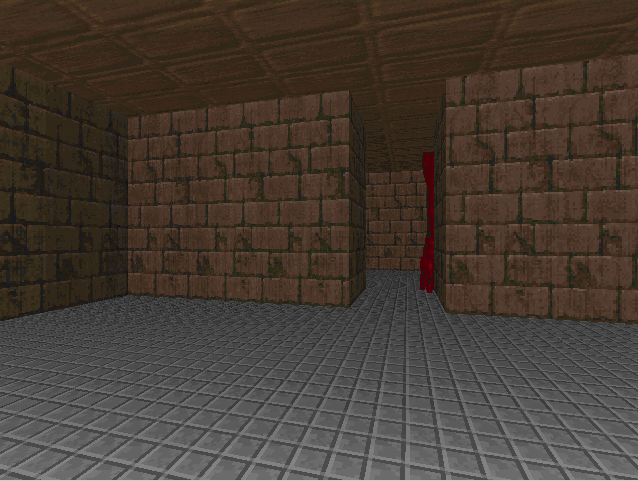}
  \end{minipage}
  \\
  \caption{Scenario map and image of agent's perspective. The goal, a red pilar, can be seen hiding behind the corner on the image on right.}
  \label{fig:Scenario_map}
\end{figure}

VizDoom allows different scenarios including different maps and sets the available buttons for the agent. Hence, we created our own scenario map (see Fig. \ref{fig:Scenario_map}) to assess the performance of the proposed work. In the scenario map, the agent's task is to navigate in the rooms and reach to the target goal. The actions of the agent are: turn left, turn right, move forward and move backward. 

The episode ends when agent reaches the goal or on timeout. The agent gets a reward of $1$ if it reaches the goal, otherwise the agent receives $-1$ reward on every game tick. We want to emphasize that this environment could be solved with recurrent neural networks like the one used in \cite{mnih2016asynchronous}. However, the goal of this work is not to solve the task, rather it is to study if audio information can help an agent to complete the task.

\subsection{Training}

The training process starts with initialization parameters such as start $\epsilon$, end $\epsilon$, discount factor and learning rate. The learning process is run for multiple training epochs where each epoch contains many learning steps. During the learning process, the value of $\epsilon$ is linearly decreased. Initially, the agent makes completely random actions (i.e., exploration). Afterwards, the actions are selected based on policy (i.e., exploitation). The last training model is used to evaluate the performance of the system. 

\subsection{Hyperparameters}
The convolutional neural network model is trained using RMSProp algorithm and minibatches of size 64. The most recent $10,000$ timesteps have been stored in the replay memory. While the discount factor is set to $0.99$, the learning rate is fixed to $0.00025$.  The baseline and proposed systems are trained for $600,000$ training steps. In the experiments where goal only spawns in the big room , the timeout was set to $1000$. In the experiment with all five rooms, timeout was set to $2100$.

Since the VizDoom environment provides a relatively high resolution image (640x480), the resolution is downsampled to $40 \times 40$ at each frame step. This reduces the computational time of training the neural network models. We have used the three channels of the image. For the audio features, we feed vectors of size $1 \times 114$ and $1 \times 100$ for the pitch and raw audio samples, respectively.  

The same CNN architectures are used both for the image and audio features. While the first layer is a convolutional layer with 8 filters of size $3 \times 3$ and a stride of $2$, the second layer is a convolutional layer with $16$ filters of size $3 \times 3$ and stride of $2$. The next fully connected layer transforms the input to $128$ units which are then transformed by another fully connected layer to an output size equal to the $4$ number of actions in each game.

An $\epsilon$ greedy policy is applied both during training and testing. During the training of DQN, $\epsilon$ starts initially with $1.0$ and gradually anneals to $0.1$ after a number of steps. During the initial training, the possible actions are uniformly selected but as the training progresses, the optimal action is selected more frequently. This allows maximum exploration at the beginning which eventually switch to exploitation. During testing, the value of $\epsilon$ is fixed to $0.05$. 

We use frame-skip of $10$ frames, i.e. the agent is given a chance to select action only every 10th game frame (corresponds to $285$ms of in-game time, as runs at $35$ frames-per-second). The previously chosen action is repeated over all frames between these agent steps.


\subsection{Performance Metric}
To assess the performance of reinforcement learning, it is common to plot the time versus reward and visualize how the reward increases/decreases during training \cite{mnih2016asynchronous}. 

In this work, we have used the average reward per episode to evaluate the performance of the baseline (no audio) and proposed work. Since the average rewards of individual runs vary considerably, it is common to launch multiple runs and plot the time versus average of these runs. Hence, the reported results in the following subsection are run and averaged over five experiment runs, i.e., each point in the learning curve figures is averaged over five episodes.

\section{Experimental Results}
The experiments are carried out using different types of setups: the target is spawned randomly in the big room  and the target is spawned randomly in any of the five rooms (see Fig.\ref {fig:Scenario_map}). The agent always starts in the center of the big room facing to random direction. 

For the experiment that spawns the target randomly in the big room, we have observed that using visual information both with pitch and raw audio provides better average reward per episode compared to using only the visual. While the use of only visual yields average reward of $-150$ over the last $100$ episodes, the augmentation of audio features to the visual yields average reward of $-100$ over the last same $100$ episodes. 

\begin{figure}[t!]
\centering
\includegraphics[width=\columnwidth]{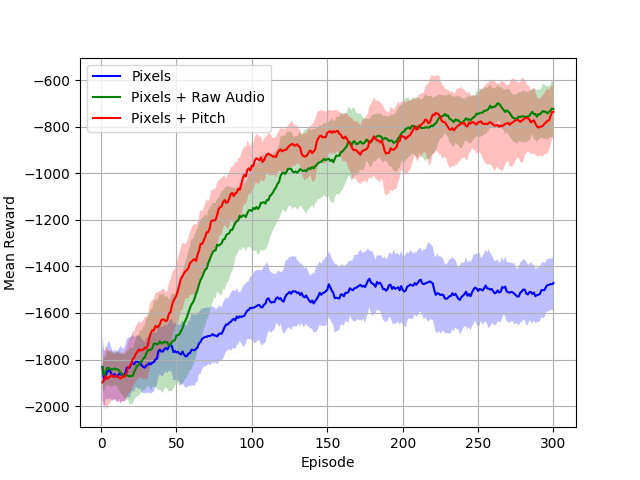}
\caption{ \it  Mean reward of using only pixels, pixels with raw audio and pixels with pitch. The target is spawned randomly in any of the five rooms for each episode. The mean reward is computed every $2000$ steps. }
\label{fig:Start_Random_Goal_Random_in_All_Rooms}
\end{figure}

We have also carried out other experiments where the target is placed randomly in any of the five rooms. From the learning curves of Fig. \ref{fig:Start_Random_Goal_Random_in_All_Rooms}, we can see that the agent does not reach the target most of the time using only visuals. But, the use of  audio features together with the visual one makes the agent to reach the goal most of the time.  The figure exhibits that while using visual and audio features provides $-750$ average reward during the last episodes, the use of only visual gives $-1500$ average reward in same episodes. Fig. \ref{fig:Start_Random_Goal_Random_in_All_Rooms} shows that the average learning curves of using visual together with audio features provide better mean rewards compared to using only the visual. We have also assessed the individual learning curves of these rewards. The individual learning curves also display that the addition of audio features to the visual ones yield better learning curve compared to the one based only on visuals.

Table \ref{table:test_results} shows the test results of $100$ episodes using visual and visual with audios. The test experiment is carried out using the last training model. The table compares the mean reward and number of steps required to reach the target using visual and audio features for the two environment setups (i.e., target randomly spawned in big room and target randomly spawned in any of the five rooms). 

From the table, we can observe that although the use of only visual enables the agent to reach the target almost all the time for the experiment that spawns the target in the big room, the addition of complementary audio features to the visual one provides lower number of steps to reach the target compared to using visual.
  
But, when we compare the average success rate and required number of steps for the experiment that spawns the target in any of the five rooms, we observe a significant improvement both in success rate and required number of steps. The use of only visual provides average success rate of $43$\%. But, the augmentation of visual with raw audio, and visual with pitch provides average success rates of $87$\% and $86$\%, respectively. Similarly, the average required number of steps to reach the target using only visual information is $1420$. But, the addition of complementary raw audio and pitch to the visual reduces the number of steps to $751$ and $614$, respectively. From the results of the table, we can see that it is difficult for the agent to make intelligent moves to reach to the target using only visual information in difficult scenarios. 

\begin{table}[h!]
  \begin{center}
    \caption{Average success rate and average number of steps of the test experiment. The test experiment is carried out on $100$ game episodes using the last training model.}
    \label{table:test_results}
    \begin{tabular}{c|l|c|r} 
    &  \textbf{Target} & \textbf{Target} & \textbf{Target}\\
     &       \textbf{randomly} & \textbf{fixed in} & \textbf{randomly}\\
      &            \textbf{in big room} & \textbf{big room} & \textbf{in all rooms}\\
      &success/ & success/ & success/ \\
      &steps & steps & steps \\
      \hline
      Visual & 99/132 & 99/164 & 38/1502\\
      Visual + Raw  & 99/98 & 99/138 & 91/588\\
      Visual + Pitch & 99/95 & 99/98 & 93/514\\
    \end{tabular}
  \end{center}
\end{table}

\begin{figure}[t!]
\centering
\includegraphics[width=\columnwidth]{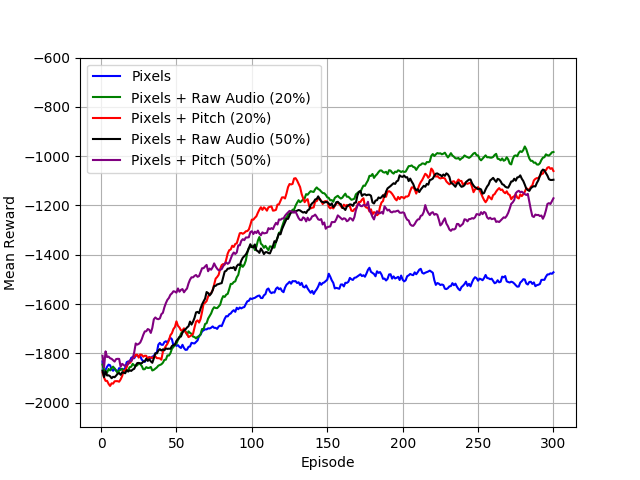}
\caption{ \it  Mean reward of using only pixels, pixels with raw audio and pixels with pitch. The audio features are occasionally used with 20\% and 50\% probability chances. The mean reward is computed every $2000$ steps. }
\label{fig:Start_Random_Goal_Random_in_All_Rooms_Sometimes}
\end{figure}

In the results of Fig. \ref{fig:Start_Random_Goal_Random_in_All_Rooms}, the audio features are always used together with the visual one to get the best action. We have also carried out other experiments where the visual is used all the time but the audio features are occasionally used. Instead of receiving audio at every time step, the agent gets audio in addition to image with certain probability. We test with probability values of [20\%, 50\%] and compare these results. Figure \ref{fig:Start_Random_Goal_Random_in_All_Rooms_Sometimes} displays the results of these experiments. We see that the use of visual and occasional use of audio features gives better average reward compared to using only the visuals. But, its reward is not as good as the one that uses audio information all the time. After conducting the experiments, we noticed that the goal spawned inside the wall (i.e., it is unreachable) $2-3\%$ of the time. However, this affects to all of the experiment and the comparison is possible.

The results of Fig. \ref{fig:Start_Random_Goal_Random_in_All_Rooms} and Fig. \ref{fig:Start_Random_Goal_Random_in_All_Rooms_Sometimes} show that providing audio features to the visual all the time provides better average reward than providing audio features occasionally.  While the provision of audio features to the visual all the time gives $-800$ reward during the last episodes, the occasional use of audio features yields average reward of $-1000$ during the same episodes.

The experimental results of the training and test demonstrate the use of audio features for reinforcement learning task. The reported results show that audio features are useful for the reinforcement learning agent in scenarios where visual information may not be found. The experimental results also demonstrate that the addition of complementary audio information to the visual one enables the agent to reach the goal faster than using only the visuals.

\section{Conclusions} \label{conclusions}
In this work, we have proposed the use of audio features for reinforcement learning and evaluate its performance in ViZDoom environment. First of all, the simulation results show that the use of visual with audio provides better average reward and lower number of steps compared to using only visual information. Secondly, the results demonstrate the usefulness of audio features for the autonomous agent in scenarios where the agent may not get visual information. The results of our work manifest the usefulness of audios for reinforcement learning (i.e., agents do benefit from hearing).

The future work could focus on confirming these results in different environments and tasks such as other video games or high-fidelity audio simulations. This could be done with or without visual information, and comparing it against visual information.

\section{Acknowledgements}
 This research was partially funded by the Academy of Finland (grant \#313970) and  Finnish Scientific Advisory Board for Defence (MATINE) project \#2500M-0106. We gratefully acknowledge the support of NVIDIA Corporation with the donation of the Titan Xp GPU used for this research.

\bibliographystyle{IEEEtran}
\bibliography{template.bbl}
\end{document}